\documentclass[10pt,twocolumn,letterpaper]{article}

\usepackage{iccv}
\usepackage{times}
\usepackage{epsfig}
\usepackage{graphicx}
\usepackage{amsmath}
\usepackage{amssymb}
\usepackage{authblk}

\usepackage[pagebackref=true,breaklinks=true,colorlinks,bookmarks=false]{hyperref}

\iccvfinalcopy 

\def\iccvPaperID{1775} 

\begin{document}

\title{Drone-based Object Counting by Spatially Regularized Regional Proposal Network}

\author[1]{Meng-Ru Hsieh}
\author[2]{Yen-Liang Lin}
\author[1]{Winston H. Hsu}
\affil[1]{National Taiwan University, Taipei, Taiwan}
\affil[2]{GE Global Research, Niskayuna, NY, USA}
\affil[ ]{\tt\small\url{mrulafi@gmail.com, yenlianglintw@gmail.com, whsu@ntu.edu.tw}}

\maketitle

\begin{abstract}
Existing counting methods often adopt regression-based approaches and cannot precisely localize the target objects, which hinders the further analysis (e.g., high-level understanding and fine-grained classification). In addition, most of prior work mainly focus on counting objects in static environments with fixed cameras. Motivated by the advent of unmanned flying vehicles (i.e., drones), we are interested in detecting and counting objects in such dynamic environments. We propose Layout Proposal Networks (LPNs) and spatial kernels to simultaneously count and localize target objects (e.g., cars) in videos recorded by the drone. Different from the conventional region proposal methods, we leverage the spatial layout information (e.g., cars often park regularly) and introduce these spatially regularized constraints into our network to improve the localization accuracy. To evaluate our counting method, we present a new large-scale car parking lot dataset (CARPK) that contains nearly 90,000 cars captured from different parking lots. To the best of our knowledge, it is the first and the largest drone view dataset that supports object counting, and provides the bounding box annotations.
\end{abstract}

\vspace{-0.5 cm}
\section{Introduction}

\begin{figure}[th]
\begin{center}
   \includegraphics[width=1\linewidth ]{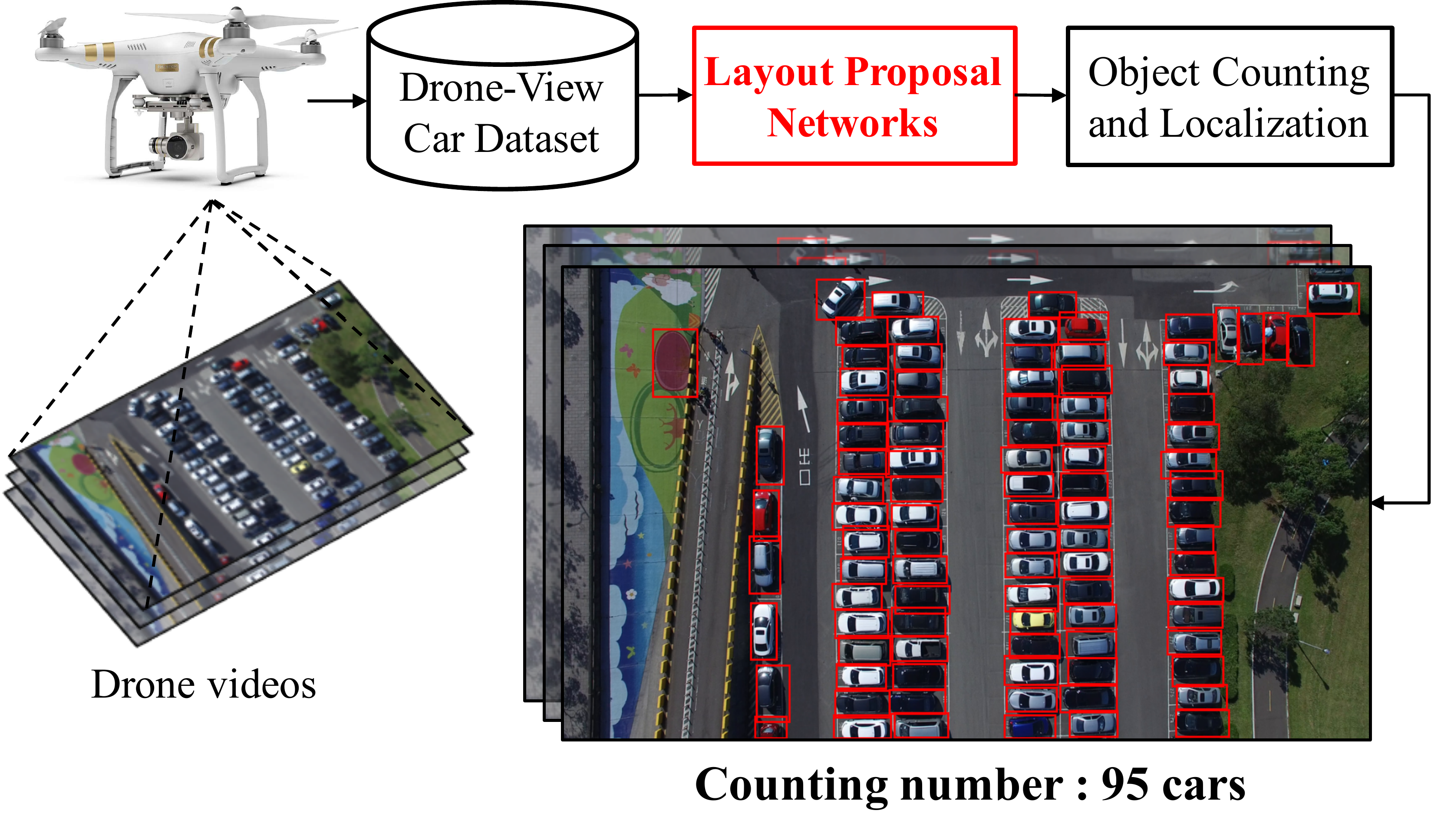}
\end{center}
   \caption{We propose a Layout Proposal Network (LPN) to localize and count objects in drone videos. We introduce the spatial constraints for learning our network to improve the localization accuracy. Detailed network structure is shown in Figure \ref{figure:LPN}.}
\label{figure:Figure1}
\end{figure}

\begin{table*}[t]
	\caption{Comparison of aerial view car-related datasets. In contrast to the PUCPR dataset, our dataset supports a counting task with bounding box annotations for all cars in a single scene. Most important of all, compared to other car datasets, our CARPK is the only dataset in drone-based scenes and also has a large enough number in order to provide sufficient training samples for deep learning models.}
\begin{center}
\begin{tabular}{|c|c|c|c|c|c|c|c|c|}
\hline
Dataset & Sensor & Multi Scenes & Resolution & Annotation Format & Car Numbers & Counting Support\\
\hline\hline
OIRDS \cite{34_tanner2009overhead} & satellite & \checkmark & low & bounding box & 180 & \checkmark \\
VEDAI \cite{35_razakarivony2016vehicle} & satellite & \checkmark & low & bounding box & 2,950 & \checkmark \\
COWC \cite{04_mundhenk2016large} & aerial & \checkmark & low & car center point & 32,716 & \checkmark \\
\hline\hline
PUCPR \cite{01_de2015pklot} & camera & $\times$ & high & bounding box & 192,216 & $\times$ \\
CARPK [ours] & drone & \checkmark & high & bounding box & 89,777 & \checkmark \\
\hline
\end{tabular}
\end{center}
\label{table:t1}
\end{table*}

\begin{figure*}[t]
\begin{center}
\begin{tabular} {c c c c c c}
	\includegraphics[width=1in]{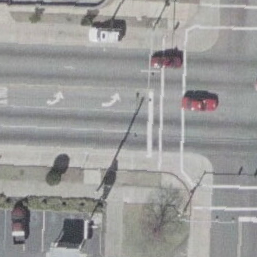} &
	\includegraphics[width=1in]{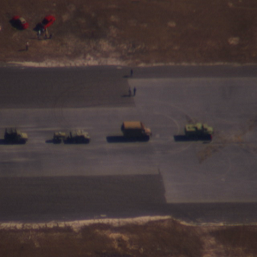} &
	\includegraphics[width=1in]{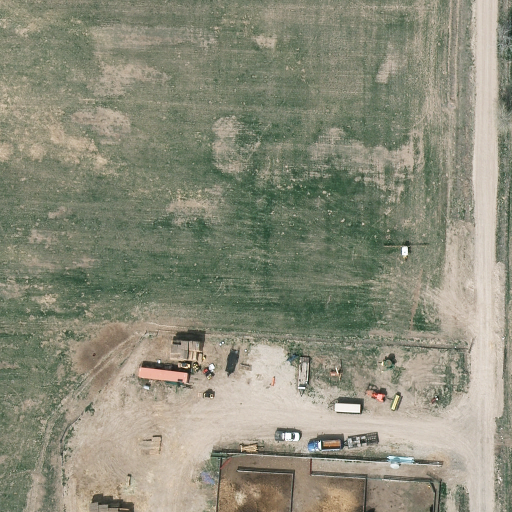} &
	\includegraphics[width=1in]{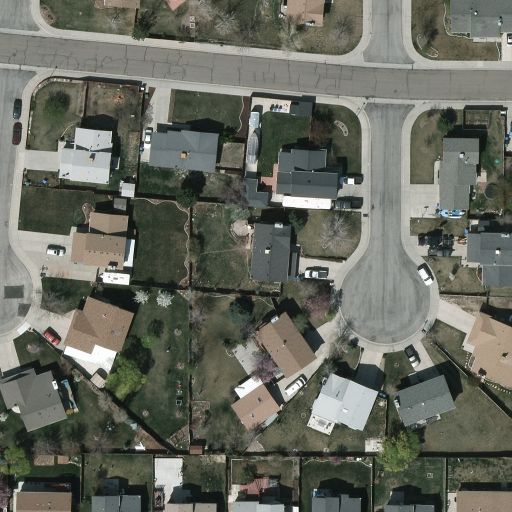} &
	\includegraphics[width=1in]{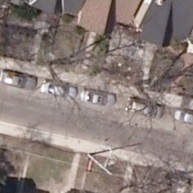} &
	\includegraphics[width=1in]{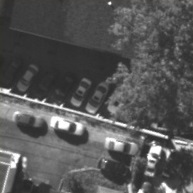} \\
\end{tabular}
\begin{tabular}{c c c}
	\makebox[5.5cm][c]{(a)} &
	\makebox[5.5cm][c]{(b)} &
	\makebox[5.5cm][c]{(c)} \\\\
\end{tabular}
\begin{tabular} {c c c c}
	\scalebox{1.58}{\includegraphics[width=1in]{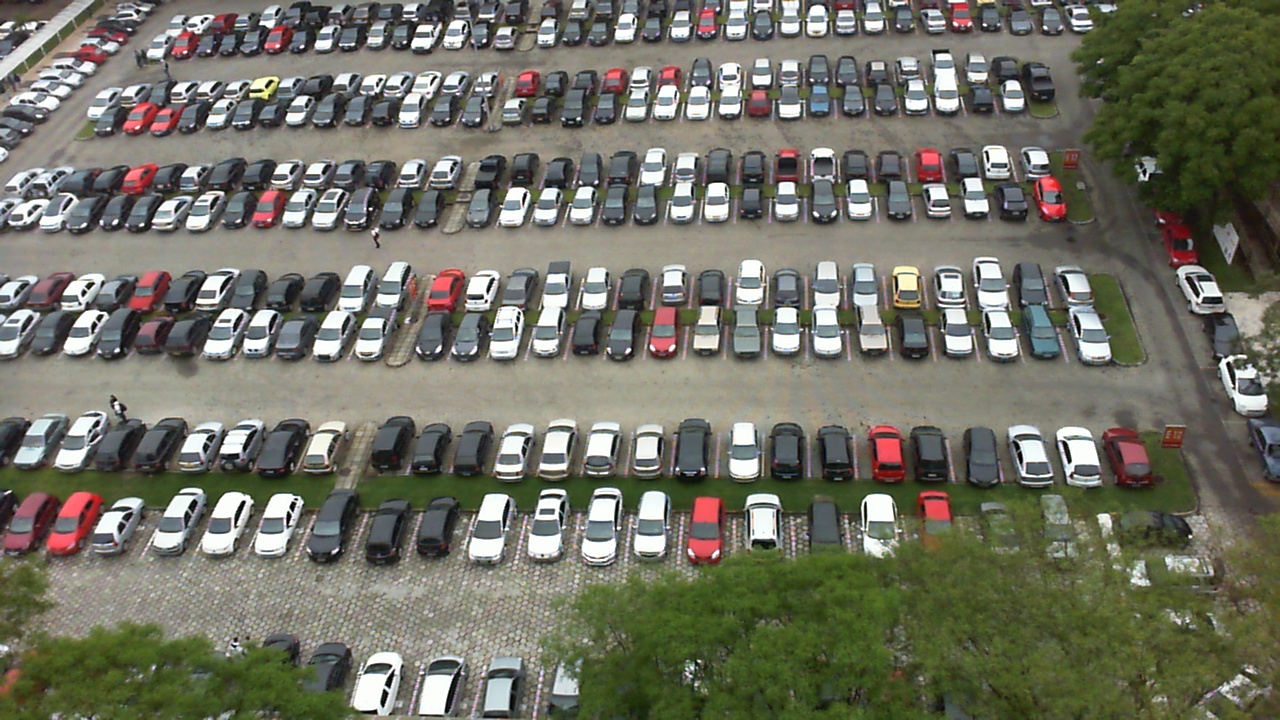}} &
	\scalebox{1.58}{\includegraphics[width=1in]{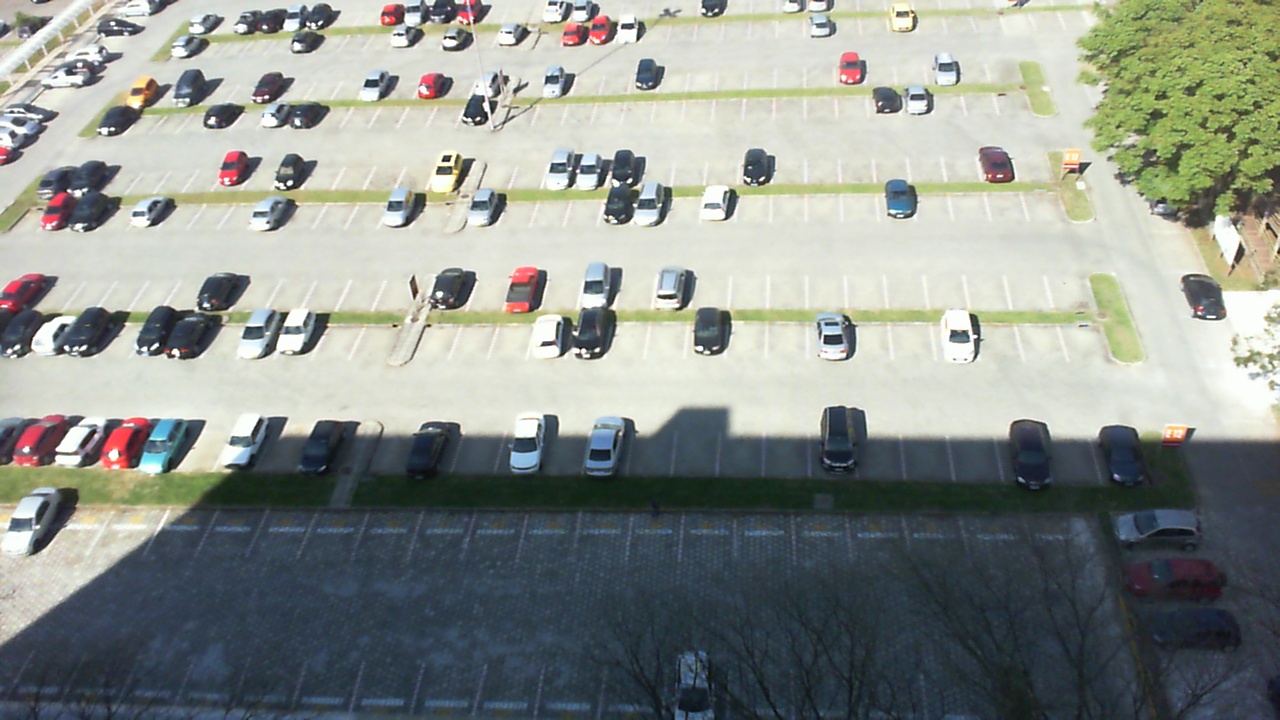}} &
	\scalebox{1.58}{\includegraphics[width=1in]{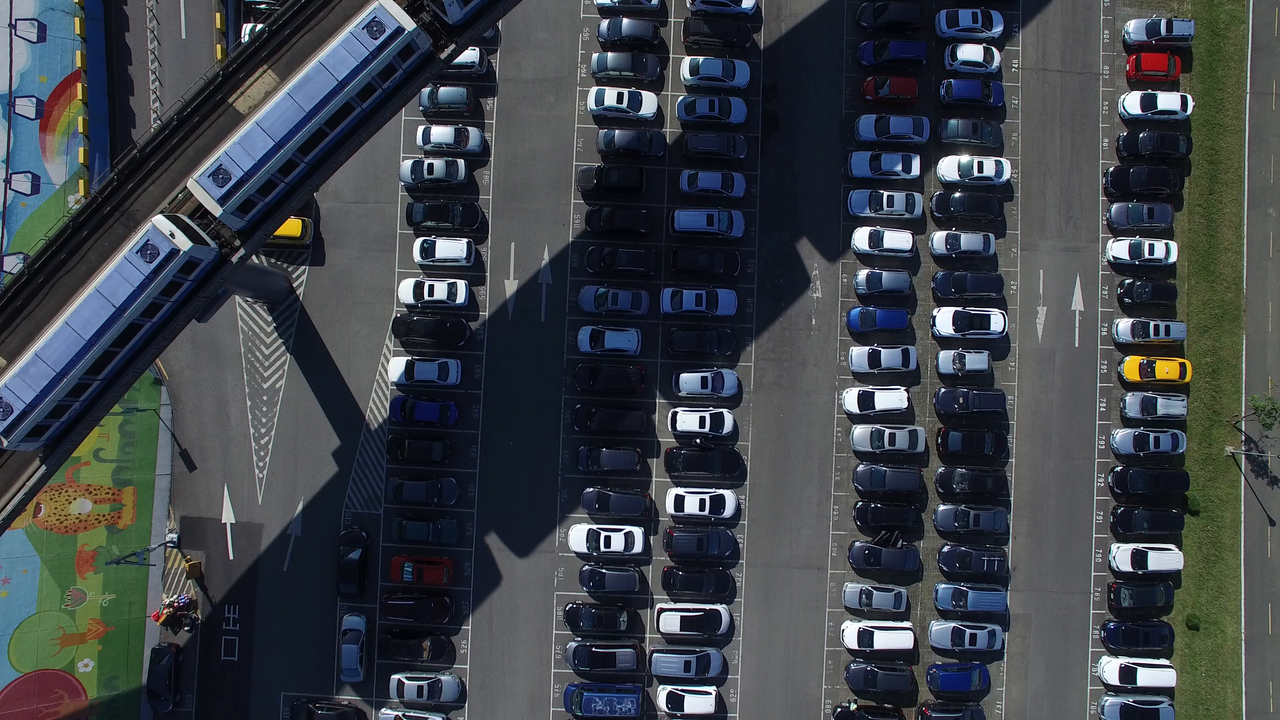}} &
	\scalebox{1.58}{\includegraphics[width=1in]{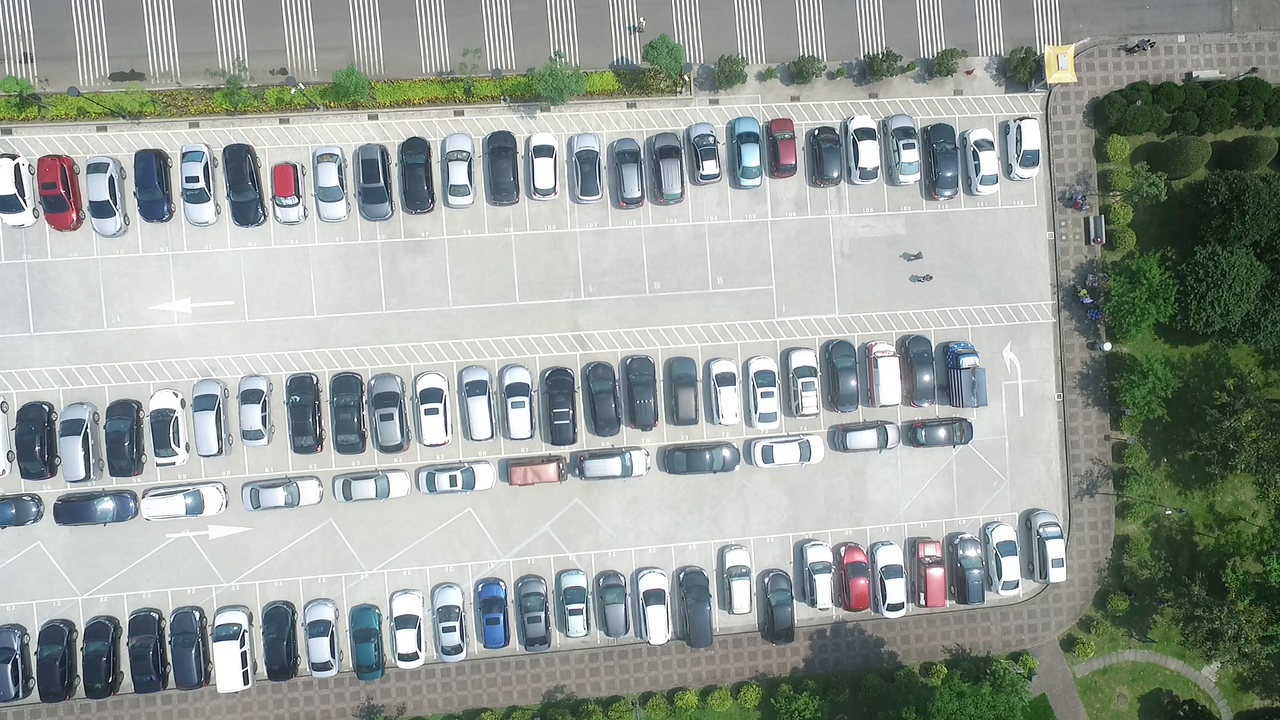}} \\
\end{tabular}
\begin{tabular}{c c c c c c}
	\makebox[0cm][c]{}    &
	\makebox[7.6cm][c]{(d)} &
	\makebox[0cm][c]{}    &
	\makebox[0cm][c]{}    &
	\makebox[7.6cm][c]{(e)} &
	\makebox[0cm][c]{} \\
\end{tabular}
\end{center}
\caption{(a), (b), (c), (d), and (e) are the example scenes of OIRDS \cite{34_tanner2009overhead}, VEDAI \cite{35_razakarivony2016vehicle}, COWC \cite{04_mundhenk2016large}, PUCPR \cite{01_de2015pklot}, and CARPK (ours) dataset respectively (two images for each dataset). Comparing to (a), (b), and (c), the PUCPR dataset and the CARPK dataset have greater number of cars in a single scene which is more appropriate for evaluating the counting task.}
\label{figure:f2}
\end{figure*}

With the advent of unmanned flying vehicles, new potential applications emerge for unconstrained images and videos analysis for aerial view cameras. In this work, we address the counting problem for evaluating the number of objects (e.g., cars) in drone-based videos. Prior methods \cite{01_de2015pklot,02_amato2016car,03_ahrnbom2016fast} for monitoring the parking lot often assume that the locations of the monitored objects of a scene are already known in advance and the cameras are fixed, and cast car counting as a classification problem, which makes conventional car counting methods not directly applicable in unconstrained drone videos.

Current object counting methods often learn a regression model that maps the high-dimensional image space into non-negative counting numbers \cite{39_wang2011automatic,04_mundhenk2016large}.
However, these methods can not generate precise object positions, which limits the further investigation and applications (e.g., recognition). 

We observe that there exists certain layout patterns for a group of object instances, which can be utilized to improve the object counting accuracy.  
For example, cars are often parked in a row and animals are gathered in a certain layout (e.g., fish torus and duck swirl). 
In this paper, we introduce a novel Layout Proposal Network (LPN) that counts and localizes objects in drone videos (Figure \ref{figure:Figure1}).
Different from existing object proposal methods, we introduce a new spatially regularized loss for learning our Layout Proposal Network. Note that our method learns the general adjacent relationship between object proposals and is not specific to a certain scene.  

Our spatially regularized loss is a weighting scheme that re-weights the importance scores for different object proposals and encourages region proposals to be placed in correct locations. 
It can also generally be embedded in any object detector system for object counting and detection. 
By exploiting spatial layout information, we improve the average recall of state-of-the-art region proposal approaches on a public PUCPR dataset \cite{01_de2015pklot} (from 59.9\% to 62.5\%). 

For evaluating the effectiveness and reliability of our approach, we introduce a new large-scale counting dataset CARPK (Table~\ref{table:t1}). 
Our dataset contains 89,777 cars, and provides bounding box annotations for each car.
Also, we consider the sub-dataset PUCPR of PKLot \cite{01_de2015pklot} which is the one that the scenes are closed to the aerial view in the PKLot dataset. Instead of a fixed camera view from a high story building (Figure~\ref{figure:f2}) in the PUCPR dataset, our new CARPK dataset provide the first and the largest-scale drone view parking lot dataset in unconstrained scenes.
Besides, the PUCPR dataset can be only used in conjunction with a classification task, which classifies the pre-cropped images (car or not car) with given locations. 
Moreover, the PUCPR dataset only annotates partial region-of-interest parking areas, and is therefore unable to support a counting task. 
Since our task is to count objects in images, we also annotate all cars in single full-image for the partial PUCPR dataset. 
The contents of our CARPK dataset are unscripted and diverse in various scenes for 4 different parking lots. 
To the best of our knowledge, our dataset is the first and the largest drone-based dataset that can support a counting task with manually labelled annotations for numerous cars in full images. 
The main contributions of this paper are summarized as follows:
 
\begin{enumerate}
\item To our knowledge, this is the first work that leverages spatial layout information for object region proposal. We improve the average recall of the state-of-the-art region proposal methods (i.e., 59.9\% \cite{21_ren2015faster} to 62.5\%) on a public PUCPR dataset.

\item We introduce a new large-scale car parking lot dataset (CARPK) that contains nearly 90,000 cars in drone-based high resolution images recorded from the diverse scenes of parking lots. 
Most important of all, compared to other parking lot datasets, our CARPK dataset is the first and the largest dataset of parking lots that can support counting\footnote{The images and annotations of our CARPK and PUCPR+ are available at https://lafi.github.io/LPN/}.

\item We provide in-depth analyses for different decision choices of our region proposal method, and demonstrate that utilizing layout information can considerably reduce the proposals and improve the counting results.
\end{enumerate}

\section{Related Work}


\subsection{Object Counting}

Most contemporary counting methods can be broadly divided into two categories. One is counting by regression method, the other is counting by detection instance \cite{38_moranduzzo2014automatic, 49_kamenetsky2015aerial}. Regression counters are usually a mapping of the high-dimension image space into non-negative counting numbers. Several methods \cite{44_an2007face, 45_chan2008privacy, 46_chen2013cumulative, 47_chen2012feature, 48_kong2006viewpoint} try to predict counts by using global regressors trained with low-level features. However, global regression methods ignore some constraints, such as the fact that people usually walk on the pavement and the size of instances. There are also a number of density regression-based methods \cite{51_rodriguez2011density, 52_lempitsky2010learning, 54_arteta2014interactive} which can estimate object counts by the density of a countable object and then aggregate over that density.

Recently, a wealth of works introduce deep learning into the crowd counting task. Instead of counting objects for constrained scenes in the preivous works, Zhang \etal \cite{53_zhang2015cross} address the problem of cross-scene crowd counting task, which is the weakness of the density estimation method in the past. Sindagi \etal \cite{55_sindagi2017generating} incorporate global and local contextual information for better estimating the crowd counts. Mundhenk \etal \cite{04_mundhenk2016large} evaluate the number of cars in a subspace of aerial imagery by extracting representations of image patches to approximate the appearance of object groups. 
Zhang \etal \cite{57_zhang2017fcn} leverage FCN and LSTM to jointly estimate the vehicle density and counts in low resolution videos from city cameras. However, the regression-based methods can not generate precise object positions, which seriously limits the further investigation and application (e.g., high-level understanding and fine-grained classification).

\subsection{Object Proposals}

Recent years have seen deep networks for region proposals developing well. Because detecting objects at several positions and scales during inference time requires a computationally demanding classifier, the best way to solve this problem is to look at a tiny subset of possible positions. A number of recent works prove that deep networks-based region proposal methods have surpassed the previous works \cite{17_uijlings2013selective, 28_arbelaez2014multiscale, 29_zitnick2014edge, 30_cheng2014bing}, which are based on the low-level cues, by a large margin.

DeepMask \cite{32_pinheiro2015learning}, which is developed for learning segmentation proposals, has, compared to Selective Search \cite{17_uijlings2013selective}, ten times fewer proposals (100 v.s. 1000) at the same performance. The state-of-the-art object proposal method, Region Proposal Networks (RPNs) \cite{21_ren2015faster}, has also shown that they just need 300 proposals and can surpass the result of 2000 proposals generated by \cite{17_uijlings2013selective}. Other works like Multibox \cite{25_szegedy2014scalable} and Deepbox \cite{29_zitnick2014edge} also have higher proposal recall with fewer number of region proposals than the previous works which are based on low-level cues. However, none of these region proposal methods have considered the spatial layout or the relation between recurring objects. Hence, we propose a Layout Proposal Networks (LPNs) that leverages thus structure information to achieve higher recall while using a smaller number of proposals.

\section{Dataset}

Since there is a lack of large standardized public datasets that contain numerous collections of cars in drone-based images, it has been difficult to create an automated counting system for deep learning models.
For instance, OIRDS \cite{34_tanner2009overhead} has merely 180 unique cars. 
The recent car-related dataset VEDAI \cite{35_razakarivony2016vehicle} has 2,950 cars, but these are still too few to utilize for the deep learners. 
A newer dataset COWC \cite{04_mundhenk2016large} has 32,716 cars, but the resolutions of images remain low. It has only 24 to 48 pixels per car. Besides, rather than labelling in the format of bounding box, the annotation format is the center pixel point of a car which can not support further investigation, such as car model retrieval, statistics of brands of car, and exploring which kind of car most people will drive in the local area.
Moreover, all above datasets are low resolution images and cannot provide detail informations for learning a fine-grained deep model. 
The problems of existing dataset are : 1) low resolution images which might harm the performance of the model trained on them and 2) less car numbers in the dataset which has the potential to cause overfitting during training a deep model.

Because existing datasets have these aforementioned problems, we have created a large-scale car parking lot dataset from drone view images, which are more appropriate to deep learning algorithms. It supports object counting, object localizing, and further investigations by providing the annotations in terms of bounding boxes. 
The most similar public dataset to ours, which also has the high resolution of car images, is the sub-dataset PUCPR of PKLot \cite{01_de2015pklot}, which provides a view from the 10th floor of a building and therefore similar to drone view images to a certain degree.
However, the PUCPR dataset can be only used in conjunction with a classification task, which classifies the pre-cropped images (car or not car) with given locations. Moreover, this dataset has only annotated a portion of cars (100 certain parking spaces) from total 331 parking spaces in a single image, making it unable to support both counting and localizing tasks. Hence, we complete the annotations for all cars in a single image from the partial PUCPR dataset, called PUCPR+ dataset, which now has nearly 17,000 cars in total. Besides the incomplete annotation problem of the PUCPR, it has a fatal issue that their camera sensors are fixed and set in the same place, making the image scene of dataset completely the same -- causing the deep learning model to encounter a dataset bias problem. 

For this reason, we introduce a brand new dataset CARPK that the contents of our dataset are unscripted and diverse in various scenes for 4 different parking lots. Our dataset also contains approximately 90,000 cars in total with the view of drone.
It is different from the view of camera from high story building in the PUCPR dataset. This is a large-scale dataset for car counting in the scenes of diverse parking lots. The image set is annotated by providing a bounding box per car. All labeled bounding boxes have been well recorded with the top-left points and the bottom-right points. Cars located on the edge of the image are included as long as the marked region can be recognized and it is sure that the instance is a car. To the best of our knowledge, our dataset is the first and the largest drone view-based parking lot dataset that can support counting with manually labeled annotations for a great amount of cars in a full-image. The details of dataset are listed in Table~\ref{table:t1} and some examples are shown in Figure~\ref{figure:f2}.

\begin{figure}[t]
\begin{center}
   \includegraphics[width=1\linewidth ]{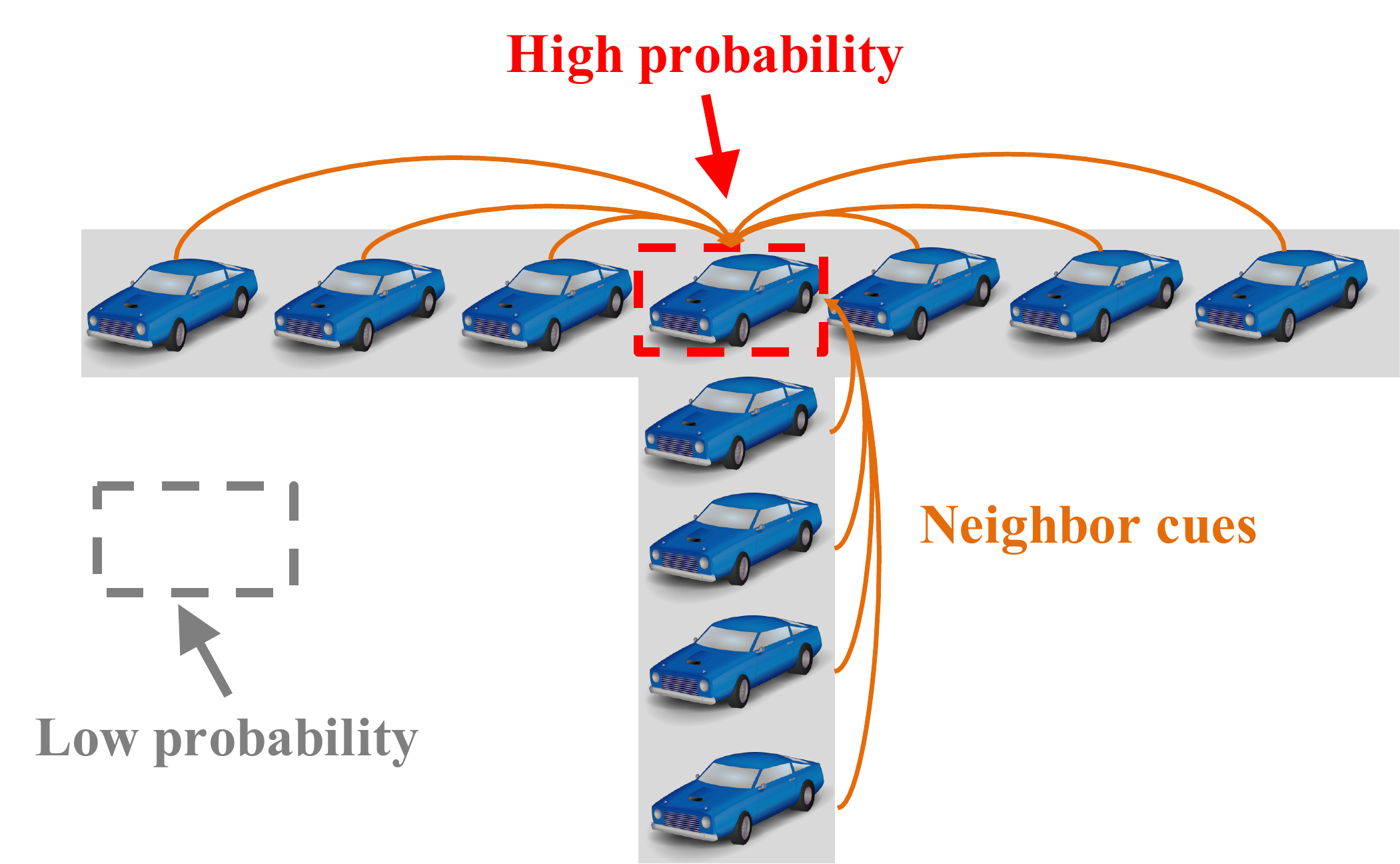}
\end{center}
   \caption{The key idea of our spatial layout scores. A predicted position which has more nearby cars can get higher confidence scores and has higher probability to be the position where the car is.}
\label{figure:SPScore}
\end{figure}

\begin{figure*}[th]
\begin{center}
   \includegraphics[width=0.75\linewidth ]{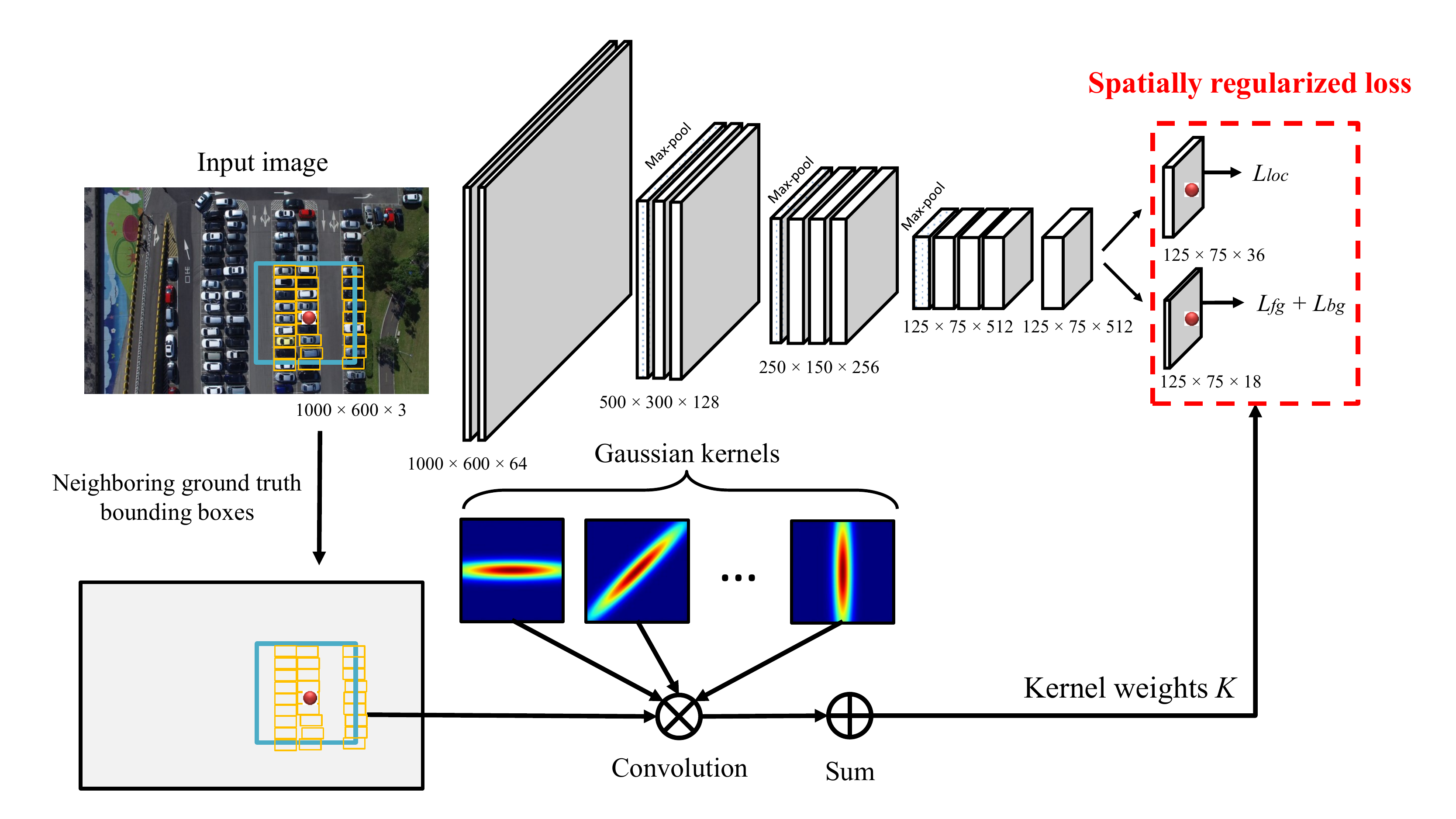}
\end{center}
   \caption{The structure of the Layout Proposal Networks.
At the loss layer, the structure weights are integrated for re-weighting the candidates to have better structure proposals. See more details in Section~\ref{sec:SPScore}.}
\label{figure:LPN}
\end{figure*}

\section{Method}

Our object counting system employs a region proposal module which takes regularized layout structure into account. It is a deep fully convolutional network that takes an image of arbitrary size as the input, and outputs the object-agnostic proposals which likely contain the instance. The entire system is a single unified framework for object counting (Figure~\ref{figure:Figure1}). By leveraging the spatial information of the object of recurring instances, LPNs module is not only concerning the possible positions but also suggesting the object detection module which direction it should look at in the image.

\subsection{Layout Proposal Network}

We observed that there exists certain layout patterns for a group of object instances, which can be used to predict objects that might appear adjacently in the same direction or near the same instances. Hence, we design a novel region proposal module that can leverage the structure layout and gather the confidence scores from nearby objects in certain directions (Figure~\ref{figure:SPScore}).

We comprehensively describe the designed network structure of LPNs (Figure~\ref{figure:LPN}) as follows. Similar to RPNs \cite{21_ren2015faster}, the network generates region proposals by sliding a small network over the shared convolutional feature map. It takes as input an $3 \times 3$ windows on last convolutional layer for reducing the representation dimensions, and then feeds features into two sibling $1 \times 1$ convolutional layers, where one is for localization and the other is for classifying whether the box belongs to foreground or background. The difference is that our loss function introduces the spatially regularized weights for the predicted boxes at each location. With the weights from spatial information, we minimize the loss of multi-task object function in networks. The loss function we use on each image is defined as:
\begin{equation} \label{eq:1}
\begin{aligned}
L(\{u_{i}\}, \{q_{i}\}, \{p_{i}\}) =& \frac{1}{N_{fg}} \sum_{i} K(c_{i}, N_{i}^{*};u_{i}^{*}) \cdot L_{fg}(u_{i}, u_{i}^{*}) \\
& + \gamma  \frac{1}{N_{bg}} \sum_{i} L_{bg}(q_{i}, q_{i}^{*}) \\
& + \lambda \frac{1}{N_{loc}} \sum_{i} L_{loc}(u_{i}^{*}, p_{i}, g_{i}^{*})
\end{aligned}
\end{equation}
where $N_{fg}$ and $N_{bg}$ are the normalized terms of the number matching default boxes for foreground and background. $N_{loc}$ is the same as $N_{fg}$ in that it only considers the number of foreground classes. The default box is marked $u_{i}^{*} = 1$ if the default box has an Intersection-over-Union (IoU) overlap higher than $0.7$ with the ground-truth box, or the default box which has the highest IoU overlap with a ground-truth box; otherwise, it is marked $q_{i}^{*} = 0$ if the IoU overlap is lower than $0.3$. The $L_{fg}(u_{i}, u_{i}^{*}) = -log[u_{i}u_{i}^{*}]$ and $L_{bg}(q_{i}, q_{i}^{*}) = -log[(1-q_{i})(1-q_{i}^{*})]$ are the negative log-likelihood that we want to minimize for true classes. Here, the $i$ is the index of predicted boxes. In front of the foreground loss, $K$ represents that we apply the spatially regularized weights for re-weighting the objective score of each predicted box. The weight is obtained by a Gaussian spatial kernel for the center position $c_{i}$ of predicted box. It will give a rearranged weight according to the $m$ neighbor ground-truth box centers, which are near to the $c_{i}$. The real neighbor centers for $c_{i}$ are denoted as $N_{i}^{*} = \{c_{1}^{*},...,c_{m}^{*}\} \in S^{c_{i}}$, which fall inside the spatial window pixels size $S$ on the input image. We use $S = 255$ in this paper to obtain a larger spatial range.  

The $L_{loc}$ is the localization loss, which is a robust loss function \cite{20_girshick2015fast}. This term is only active for foreground predicted boxes ($u_{i}^{*} = 1$), otherwise $0$. Similar to \cite{21_ren2015faster}, we calculate the loss of offsets between the foreground predicted box $p_{i}$ and the ground truth box $g_{i}$ with their center position ($x, y$), width ($w$), and height ($h$) based on the default box ($d$).
\begin{equation} \label{eq:Llos}
\begin{aligned}
L_{loc}(u_{i}^{*}, p_{i}, g_{i}^{*}) =& \sum_{i \in fg} \sum_{v \in \{x, y, w, h\}} u_{i}^{*}smooth_{L1}(p_{i}^{v}, g_{i}^{v*})
\end{aligned}
\end{equation}
, where $g_{i}^{v*}$ (similar to $p_{i}^{v}$) is defined as below:
\begin{align}
  \begin{aligned}
   g_{i}^{x*} = (g_{i}^{x} - d_{i}^{x}) / d_{i}^{w}, \\       g_{i}^{w*} = log(g_{i}^{w} / d_{i}^{w}),
  \end{aligned}
  &&
  \begin{aligned}
   g_{i}^{y*} = (g_{i}^{y} - d_{i}^{y}) / d_{i}^{h} \\       g_{i}^{h*} = log(g_{i}^{h} / d_{i}^{h})
  \end{aligned}
 \end{align}
In our experiment, we set $\gamma$ and $\lambda$ to be 1. Besides, in order to handle the small objects, instead of conv5-3 layer, we select conv4-3 layer features for obtaining better tiling default box stride on the input image and choose default box sizes approximately four times smaller $(16\times16, 40\times40, 100\times100)$ than the default setting $(128\times128, 256\times256, 512\times512)$.

\subsection{Spatial Pattern Score} \label{sec:SPScore}
Most of the objects of an instance exhibit a certain pattern between each other. For instance, cars will align in one direction on a parking lot and ships will hug the shore regularly. Even in biology, we can also find collective animal behavior that makes them look into a certain layout (e.g., fish torus, duck swirl, and ant mill). Hence, we introduce a method for re-weighting the region proposals in the training phase in an end-to-end manner. The proposed method can reduce the number of proposals in the inference phase for abating the computational cost of the counting and detection processes. It is especially important on embedded devices, such as the drone, to lower power consumption as the battery power only can provide the drone with energy to fly a mere 20 minutes.

For designing the pattern of layout, we apply different direction 2D Gaussian spatial kernels $K$ (see Eq.~\ref{eq:1}) on the space of input images, where the center of the Gaussian kernel is the predicted box position $c_{i}$. We compute the confidence weights over all positive predicted boxes. By incorporating the prior knowledge of layout from ground-truth, we can learn the weight for each predicted box. In Eq.~\ref{eq:2}, it illustrates that the spatial pattern score for predicted position $c_{i}$ is a summation of weights by the ground truth positions which are inside the $S^{c_{i}}$. We compute the score over the input triples ($c_{i}, N_{i}^{*}, u_{i}^{*}$):
 	
\begin{equation} \label{eq:2}
K(c_{i}, N_{i}^{*}, u_{i}^{*}) = \begin{cases} \sum_{\theta \in D} \sum_{j \in N_{i}^{*}}^{m} G(j ; \theta) & \mbox{if } u_{i}^{*}\mbox{ = 1} \\ 
1 & \mbox{otherwise,} \end{cases} 
\end{equation}
in which
\begin{equation} \label{eq:3}
G(j ; \theta) = \alpha \cdot e^{-(\frac{x_{j}^{\theta}}{2 \sigma_{x}^{2}}+\frac{y_{j}^{\theta}}{2 \sigma_{y}^{2}})}, 
\end{equation}
is the 2D Gaussian spatial kernel that takes different rotated radius $D = \{\theta_{1},...\theta_{r}\}$, where we use $r = 4$ ranged from 0 to $\pi$. The coordinate tuple $(x_{j}, y_{j})$ is the center position of $jth$ ground-truth box in Eq. \ref{eq:3}, and the coefficient $\alpha$ is the amplitude of the Gaussian function. All experiments use $\alpha = 1$.

We only give the weights for the foreground predicted box $c_{i}$ where it is marked $u_{i}^{*} = 1$. By the means of aggregating weights from ground-truth boxes $N_{i}^{*}$ in different direction kernels (Eq. \ref{eq:2}), we can compute a summation of scores for taking various layout structures into account. It will give a higher probability to the object position, which has larger weight. Namely, the more similar objects of instances surrounding it, the more possible the predicted boxes are the same category of instances. Therefore, the predicted box collects the confidence from the same objects which are nearby (Figure~\ref{figure:SPScore}). By leveraging spatially regularized weights, we can learn a model for generating the region proposals where the objects of instance will appear with their own layout.

\section{Experiment}
In this section, we evaluate our approach on two different datasets. The PUCPR+ dataset, made from the sub-dataset of the public PKLot dataset \cite{01_de2015pklot}, and the CARPK dataset are both used to estimate the validation of our proposed LPNs. Then, we evaluate our object counting model, which leverages the structure information on the PUCPR+ dataset and our CARPK dataset.

\subsection{Experiment Setup}
We implement our model on Caffe \cite{40_jia2014caffe}. For fairness in analyzing the effectiveness between different baseline methods, we implemented all of them based on the VGG-16 networks \cite{06_simonyan2014very} which contains 13 convolutional layers and 3 fully-connected layers. All the layer parameters of baselines and our proposed model are using the weights pre-trained on ImageNet 2012 \cite{41_russakovsky2015imagenet}, followed by fine-tuning the models on our CARPK dataset or the PUCPR+ dataset depending on the experiments. We run our experiments on the environment of Linux workstation with Intel Xeon E5-2650 v3 2.3 GHz CPU, 128 GB memory, and one NVIDIA Tesla K80 GPU. Our multi-task joint training scheme takes approximately one day to converge.

\subsection{Evaluation of Region Proposal Methods}

For evaluating the performance of our method LPNs, we use five-fold cross-validation on the PUCPR+ dataset to ensure that the same image would not appear across both training set and testing set. In order to better evaluate the recall while estimating localization accuracy, rather than reporting recall at particular IoU thresholds, we report the Average Recall (AR). It is an average of recall with IoU threshold $t$ between 0.5 to 1, where $AR = \frac{1}{t} \sum_{i}^{t} Recall(IoU_{t})$. As the metric of recall at IoU of 0.5 is not predictive of detection accuracy, proposals with high recall but at low overlap are not effective for detection \cite{42_hosang2016makes}. Therefore, adopting the IoU range of the AR metric can simultaneously measure both proposal recall and localization accuracy to better predict the result of counting and localizing performance.

\begin{table}[h]
	\caption{Result on the PUCPR+ \cite{01_de2015pklot} dataset for average recall at 100, 300, 500, 700, and 1000 proposals with the different components of approaches. The method in the middle column represents the RPN training with the small default box size on conv4-3 layer.}
\begin{center}
\begin{tabular}{|c|c|c|c|}
\hline
\#Proposals & RPN \cite{21_ren2015faster} & RPN+small & LPN (ours) \\
\hline\hline
100  & 3.2\% & 20.5\% & 23.1\% \\
\hline
300  & 9.1\% & 43.2\% & 49.3\% \\
\hline
500  & 13.9\% & 53.4\% & 57.9\% \\
\hline
700  & 17.4\% & 57.3\% & 60.7\% \\
\hline
1000 & 21.2\% & 59.9\% & 62.5\% \\
\hline
\end{tabular}
\end{center}
\label{table:t2}
\end{table}
\vspace{-0.5 cm}

We compare our proposal method LPNs against the state-of-the-art object proposal generator RPNs \cite{21_ren2015faster} on the PUCPR+ dataset with different number of the object proposals. Our results are shown in Table~\ref{table:t2}. It reveals that our proposal method LPNs, which leverages the regularized layout information, can achieve higher recall and surpass RPNs in the same number of proposals. The state-of-the-art object proposal RPNs suffer from poor performance in average recall. We refer that the factors, which affect the performance, are upon on the inappropriate anchor size and the resolution of convolutional layer features. Hence, in the same manner, we apply the smaller anchor box size on RPNs on the conv4-3 layer, which is in the same setting as our approach. Table~\ref{table:t2} shows that the RPNs utilize the small anchor size and the higher resolution feature map bring about a better improvement. It implies that the CNN model is not as powerful in scale variance as we thought when using inappropriate layers or unsuitable default box size for prediction.
This experiment also shows that the performance of our proposed model LPNs with spatial regularized constraints still outperforms the revised RPNs (e.g., 14.1\% better in 300 proposals and 8.42\% better in 500 proposals).
Besides, we also found that our method with spatial regularizer significantly performs better in the dense case \footnote{We additionally divide the PUCPR+ dataset into dense and less dense cases. Our method has 16.30\% large relative improvement compared to RPN-small in dense case, which is better than 8.27\% for less dense case. Moreover, our method localizes the bounding box more precisely, i.e., our method achieves 64.4\% recall in IoU at 0.7 which is almost 10\% better than RPN-small 54.7\% for 300 proposals.}. The result indicates that the prior layout knowledge could potentially benefit the outcome by giving the correct confidence score to the position of instances in images.

\begin{table}[h]
	\caption{Results on the CARPK dataset with different components.}
\begin{center}
\begin{tabular}{|c|c|c|c|}
\hline
\#Proposals & RPN \cite{21_ren2015faster} & RPN+small & LPN (ours) \\
\hline\hline
100  & 11.4\% & 31.1\% & 34.7\% \\
\hline
300  & 27.9\% & 46.5\% & 51.2\% \\
\hline
500  & 34.3\% & 50.0\% & 54.5\% \\
\hline
700  & 37.4\% & 51.8\% & 56.1\% \\
\hline
1000 & 39.2\% & 53.4\% & 57.5\% \\
\hline
\end{tabular}
\end{center}
\label{table:t3}
\end{table}
\vspace{-0.5 cm}

For looking into the details of the effectiveness of our approach in region proposal, we also conduct the experiment on our CARPK dataset. In order to ensure that the same or the similar image scenes would not appear across both training and testing set, which would affect the observation of validation, we take 3 different scenes of the parking lot as training set and the remaining one scene of the parking lot as testing set. Table~\ref{table:t3} reports the average recall of our methods, the state-of-the-art region proposal method RPNs, and the revised RPNs on CARPK dataset. In the experiment results, it comes as no surprise that by incorporating the additional layout prior information, our LPNs model boots both recall and localization accuracy of proposal method. Again, this result shows that the proposals generated by our approach are more effective and reliable.

\begin{figure*}[t]
\begin{center}
   \includegraphics[width=1\linewidth ]{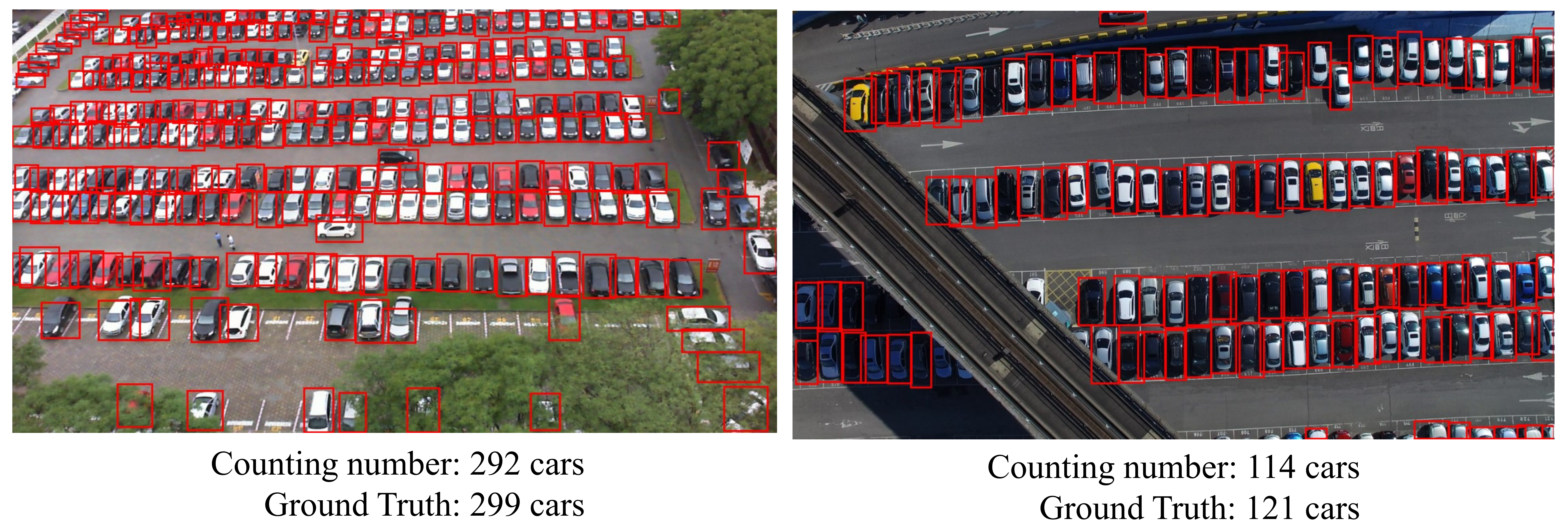}
\end{center}
   \caption{Selected examples of car counting and localizing results on the PUCPR+ dataset (left) and the CARPK dataset (right). The counting model uses our proposed LPN trained on a VGG-16 model and combined with an object detector (Fast R-CNN), where the parameters setting of confidence score is 0.5 and non maximum suppression (NMS) is 0.3 for 2000 proposals.}
\label{figure:Examples}
\end{figure*}

\subsection{Evaluation of Car Counting Accuracy}

Since the goal of our proposed approach is to count the number of cars from drone view scenes, we compare our car counting system with three state-of-the-art methods which can also achieve the car counting task. A one-look regression-based counting method \cite{04_mundhenk2016large} which is the up-to-date method for estimating the number of cars in density object counting measure and two prominent object detection systems, Faster R-CNN \cite{21_ren2015faster} and YOLO \cite{43_redmon2016you}, which have remarkable success in object detection task in recent years.

For fair comparison, all the methods are built based on a VGG-16 network \cite{06_simonyan2014very}. The only difference is that \cite{04_mundhenk2016large} uses a softmax with 64 outputs as they assumed that the maximum number of cars in a scene is sufficiently small. However, the maximum number of cars in the CARPK dataset is far more than 64. The maximum number of cars is 331 in a single scene of the PUCPR+ dataset and 188 in a single scene of the CARPK dataset. Hence, we follow the setting from \cite{04_mundhenk2016large} and train the network with a different output number to make this regression-based method compatible with the two datasets. We set the softmax to 400 outputs for the PUCPR+ dataset and 200 outputs for the CARPK dataset for evaluation. Last, the setting of two datasets, PUCPR+ and CARPK, are the same as the experiment of region proposal phase.

We employ two metrics, Mean Absolute Error (MAE) and Root Mean Squared Error (RMSE), for evaluating the performance of counting methods. These two metrics have the similar physical meaning that estimates the error between the ground-truth car numbers $y_{i}$ and the predicted car numbers $f_{i}$. MAE is the average absolute difference between ground-truth quantity $y_{i}$ and predicted quantity $f_{i}$ over all testing scenes where $MAE = \frac{1}{n} \sum_{i}^{n} |f_{i} - y_{i}|$. Similar, RMSE is the square root of the average of squared differences between ground-truth quantity and predicted quantity over all testing scenes where $RMSE =\sqrt{\frac{1}{n} \sum_{i}^{n} (f_{i} - y_{i})^{2}}$. The difference of the two metrics is that the RMSE should be more useful when large errors are particularly undesirable. Since the errors are squared before they are averaged, the RMSE gives a relatively high weight to large errors. In the counting task, these metrics have good physical meaning for representing the average counting error of cars in the scene.

\begin{table}[h]
	\caption{Comparison with the object detection methods and the global regression method for car counting on the PUCPR+ dataset. $N_{p}$ is the number of candidate boxes used in the object detector, which parameterizes the region proposal method. The "$*$" in front of the baseline methods represents that the method has been fine-tuned on PUCPR+ dataset. The "$\dagger$" represents that the method is revised to fit our dataset.}
	\vspace{-0.5 cm}
\begin{center}
\begin{tabular}{|c|c|c|c|}
\hline
Method & $N_{p}$ & MAE  & RMSE \\
\hline\hline
YOLO \cite{43_redmon2016you}		  & - & 156.72 & 200.54 \\
\hline
Faster R-CNN \cite{21_ren2015faster}  & 400 & 156.76 & 200.59 \\
\hline\hline
*YOLO 		  						  & - & 156.00 & 200.42 \\
\hline
*Faster R-CNN  						  & 400 & 111.40 & 149.35 \\
\hline
*Faster R-CNN (RPN-small)  			  & 400 & 39.88 &  47.67\\
\hline
$\dagger$One-Look Regression \cite{04_mundhenk2016large}  & - & \textbf{21.88} & 36.73 \\
\hline
Our Car Counting CNN Model 			  & 400 & 22.76 & \textbf{34.46} \\
\hline
\end{tabular}
\end{center}
\label{table:t4}
\end{table}
\vspace{-0.5 cm}

We compare three methods on the PUCPR+ dataset, where the maximum number of cars is 311 in a single scene. Since the softmax output number of \cite{04_mundhenk2016large} is designed to be 400 for the PUCPR+ dataset, we also impartially compare to this dense object counting method with the number of region proposals limited to 400, which is a strict condition to our object counter. For YOLO, we select the parameter of confidence threshold at 0.15, which gives the best performance in our dataset. 

The experimental results are shown in Table~\ref{table:t4}. The asterisk "$*$" in front of the YOLO and Faster R-CNN methods represents that the models have been fine-tuned on the PUCPR+ dataset, otherwise they are fine-tuned on the benchmark datasets (PASCAL VOC dataset and MS COCO dataset respectively), where they also have the car categories. 
Our proposed method outperforms the best RMSE on large-scale car counting, even with a very tough setting in the number of proposals. Note that we have comparable MAE performance to the state-of-the-art car counting regression method \cite{04_mundhenk2016large}, but the better RMSE implies that our method has better capability in some extreme cases. 
The methods that are fine-tuned on PASCAL and MS COCO get worse results. It reveals that the inferior outcomes are caused by the different perspective view of the object even when training with car category samples. The experiment results show that by incorporating the spatially regularized information, our Car Counting CNN model boosts the performance of counting. A counting and localization example result is shown in Figure~\ref{figure:Examples} (left).

We further compare the counting methods on our challenging large-scale CARPK dataset where the maximum number of cars is 188 in a single scene. However, different from the PUCPR+ dataset which only has one parking lot, our CARPK dataset provides various scenes of diverse parking lots for cross-scene evaluation. In the setting of \cite{04_mundhenk2016large} method, we also deign a 200 softmax output network for the CARPK dataset. In order to fairly compare the counting methods, we again restrict the number of proposals of object counter which has utilized the region proposal method with a tough number 200 \footnote{Our method gets better performance when using bigger number of proposals (e.g., 8.04 and 12.06 for 1000 proposals in MAE and RMSE respectively) in the PUCPR+ dataset. In the CARPK dataset, our method also has 13.72 and 21.77 for 1000 proposals in MAE and RMSE respectively.}. The quantitative results of car counting on our dataset are reported in Table~\ref{table:t5}. The experiment results show that our car counting approach is reliably effective and has the best MAE and RMSE even in the cross-scene estimation. An counting and localizing example result is shown in Figure~\ref{figure:Examples} (right). Still our method can generate the feasible proposals and obtain the reasonable counting result close to the real number of cars in the scenes of the parking lots.

\begin{table}[h]
	\caption{Comparison results on the CARPK dataset. The notation definition is similar to Table~\ref{table:t4}.}
	\vspace{-0.5 cm}
\begin{center}
\begin{tabular}{|c|c|c|c|}
\hline

Method & $N_{p}$ & MAE  & RMSE \\
\hline\hline
YOLO \cite{43_redmon2016you}		  & - & 102.89 & 110.02 \\
\hline
Faster R-CNN \cite{21_ren2015faster}  & 200 & 103.48 & 110.64 \\
\hline\hline
*YOLO 		 						  & - & 48.89 & 57.55 \\
\hline
*Faster R-CNN  						  & 200 & 47.45 & 57.39 \\
\hline
*Faster R-CNN (RPN-small)  			  & 200 & 24.32 &  37.62\\
\hline
$\dagger$One-Look Regression \cite{04_mundhenk2016large}  & - & 59.46 & 66.84 \\
\hline
Our Car Counting CNN Model 			  & 200 & \textbf{23.80} & \textbf{36.79} \\
\hline
\end{tabular}
\end{center}
\label{table:t5}
\end{table}
\vspace{-0.5 cm}

\section{Conclusions}
We have created the to-date largest drone view dataset, called CARPK. It is a challenging dataset for various scenes of parking lots in a large-scale car counting task. Also, in the paper, we introduced a new way for generating the feasible region proposals, which leverage the spatial layout information for an object counting task with regularized structures. The learned deep model can specifically count objects better with the prior knowledge of object layout patterns. Our future work will involve global information, such as context, road scene, and other objects which can help distinguish between false car-like instances and real cars.

\section{Acknowledgement}
This work was supported in part by the Ministry of Science and Technology, Taiwan, under Grant MOST 104-2622-8-002-002 and MOST 105-2218-E-002-032, and in part by MediaTek Inc, and grants from NVIDIA and the NVIDIA DGX-1 AI Supercomputer.


{\small
\bibliographystyle{ieee}
\bibliography{egbib}
}

\end{document}